# A New Loss Function for CNN Classifier Based on Pre-defined Evenly-Distributed Class Centroids

Qiuyu Zhu, Pengju Zhang, Xin Ye

School of Communication and Information Engineering, Shanghai University, Shanghai, CHINA

zhuqiuyu@staff.shu.edu.cn

**Abstract:** With the development of convolutional neural networks (CNNs) in recent years, the network structure has become more and more complex and varied, and has achieved very good results in pattern recognition, image classification, object detection and tracking. For CNNs used for image classification, in addition to the network structure, more and more research is now focusing on the improvement of the loss function, so as to enlarge the inter-class feature differences, and reduce the intra-class feature variations as soon as possible. Besides the traditional Softmax, typical loss functions include L-Softmax, AM-Softmax, ArcFace, and Center loss, etc.

Based on the concept of predefined evenly-distributed class centroids (PEDCC) in CSAE network, this paper proposes a PEDCC-based loss function called PEDCC-Loss, which can make the inter-class distance maximal and intra-class distance small enough in hidden feature space. Multiple experiments on image classification and face recognition have proved that our method achieve the best recognition accuracy, and network training is stable and easy to converge. Code is available in https://github.com/ZLeopard/PEDCC-Loss

## 1. Introduction

In the past few years, convolutional neural networks (CNNs) have brought excellent performance in many areas such as image classification, object detection, and face recognition. CNNs extract features from complex datasets through kinds of convolutional layers and pooling layers, and then linear layer is performed for classification. Due to the powerful feature expression and learning ability of CNNs, we can solve a variety of visual recognition tasks.

In order to addressing the drawbacks currently faced by CNNs, many researchers have proposed very effective solutions, such as data augmentation, regularization, dropout, batch normalization and various activation functions. The development of the network structure is also very rapid, from the beginning of AlexNet [1] to VGGNet [2], and to the deeper ResNet [3], ResNext [4], DenseNet [5] and SEResNet [6], etc. The advantages of CNNs are constantly magnified.

Recent research has slowly extended to the design of loss function to obtain a more distinguishing feature distribution, which means the compactness of intra-class and the discreteness of inter-class as soon as possible. Due to the strong fitting ability of the CNNs, these methods can work well and the accuracy of classification can be improved. So, more and more researchers have optimized the loss function. Due to the advantages of clear theory, easy training, and good performance, the traditional cross entropy loss function is widely used in image classification. But it is not guaranteed to obtain the optimized feature distribution mentioned above. The contrastive loss [7] and triplet loss [8] were proposed to increase the constraint on features. It can easily train large-scale data sets without being limited by display storage. But the disadvantage is that much attention is paid to local feature, leading to training difficulties and long convergence time. L-

Softmax [9] introduces the margin parameter and modifies the original Softmax function decision boundary, which increases the learning difficulty by modifies $\|W\|\|X\|\cos\theta$ to $\|W\|\|X\|\cos m\theta$, alleviating the over-fitting problem, and producing the decision margin to make the distribution more discriminative. AM-Softmax [10] set $\|W\| = \|X\| = 1$, and normalize the last layer weights and output features to reduce the difference of image's resolution in the dataset, and the impact of quantitative differences. Then, the Euclidean feature space are converted into the cosine feature space and $\cos(m\theta)$ are changed to $\cos\theta - m$, which makes the backpropagation easier. The core of Center Loss [11] is that in each batch, a class center is calculated, and the distance between each sample and the class center are minimized. Then, the mean square error combined with the Cross-Entropy loss is proposed, in which the class centers are also trained by stochastic gradient descent. However, the distances of familiar classes are not well separated, and the distribution in the Euclidean space is not uniform. For example, the class centers of class '0' and '6' in MNIST [12] are relatively closer.

In this paper, PEDCC proposed in CSAE (Zhu qiuyu *et al.*, 2019) [13] is used to generate the class centroids of the evenly distributed normalized weight, which is called PEDCC weights. We replace the weight of the classification linear layer with PEDCC weights in CNNs, and the PEDCC weights are solidified during training to maximize the inter-class distance. Thus, we get a new loss function called PEDCC-loss by apply the fixed PEDCC weight to AM-Softmax. In the same time, we add a constrain condition similar to Center Loss [11] to calculated the mean square error loss (MSE loss) between the sample feature and PEDCC centroids. This can optimize the feature embedding to enforce higher similarity for intra-class samples and diversity for inter-class samples. The two losses sum for the backpropagation to update the parameter weights before the classification linear layer. Compared with Center loss [11], the class centroid is fixed and uniform, and PEDCC weights are applied to AM-Softmax loss [10]. The method makes the feature distribution optimal for the compactness of intra-class and the discreteness of inter-class.

The overall system diagram is shown in Figure 1. Details of the proposed method are given in Section 3.

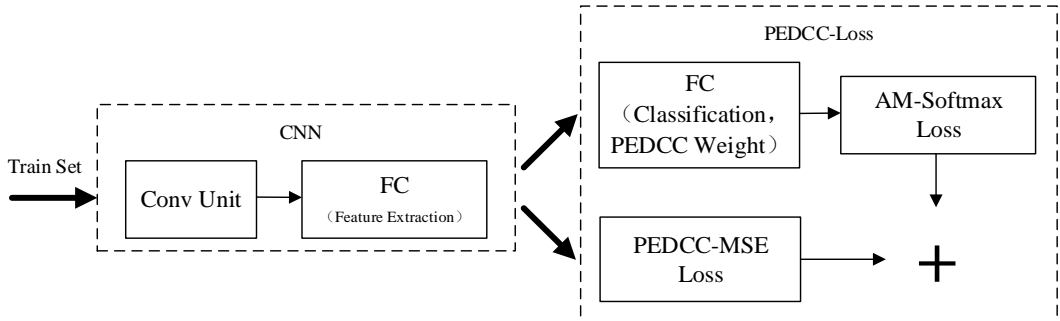

Figure 1. The PEDCC-loss

Our main contributions are as follows:

(1) The PEDCC introduce by CSAE [13] is used as weight parameter to solidify the classification layer in the convolutional neural network, which also reduces the parameter number of the whole network and improves the training speed and accuracy.

(2) PEDCC weights are applied to AM-Softmax [10] loss, and the improved MSE loss between the feature vector and the predefined class centroid are calculated. Two losses are added to form PEDCC-Loss. In the final stage of training, an optional finetuning trick are adopted to further improve the accuracy of classification.

(3) For the image recognition and face recognition tasks, multiple datasets (EMNIST [14], CIFAR100 [15], FaceScrub [16] and LFW [17]) are evaluated. Compared with the latest research work, our method achieves the optimal recognition accuracy, and network training is stable and easy to converge.

## 2. Related Work

There are various loss functions in CNNs. Traditional loss functions include Hinge loss, The contrastive loss [7], Triplet loss [8], and the most commonly used Softmax loss function. But the Softmax loss is not good at reducing the intra-class variation. To address this problem, L-Softmax [9] introduce the margin parameter to multiply the angle between the classes in order to increase learning difficulty. However, due to the $\cos(m\theta)$, the training is difficult to converge. A-Softmax [18] introduced an conceptually appealing angular margin to push the classification boundary closer to the weight vector of each class. AM-Softmax [10] and CosFace [19] are also directly adds cosine margin penalty to the target logit, which obtains better performance compared to A-Softmax [18], and easier to implement and converge. ArcFace [20] moved cosine margin to the angular margin by changing $\cos\theta - m$ to $\cos(\theta + m)$, and also discuss the impact of different decision boundaries. But it is found by our experiments that there is no good universality, and it cannot be applied to different classification tasks. Center loss [11] innovatively discussed the distance between each sample and the class center. The mean square error combined with the cross entropy was added to compress the intra-class distance. however, the center of each class is continuously optimized during the training process.

Let's review the Softmax loss and define the $i$th input feature as $x_i$ and label $y_i$. It can be expressed as follows:

$$L_{softmax} = \frac{1}{N}\sum_i L_i = \frac{1}{N}\sum_i -log\left(\frac{e^{f_{y_i}}}{\sum_j e^{f_j}}\right) \quad (1)$$

where $y_j$ represents the $j$th element of the class output vector of the final fully connected layer, and N is the number of training samples. Since $f_{y_i}$ can be expressed as $f_{yi} = W_{yi}^T X_i$, and the final loss function can be written as:

$$L_i = -log\left(\frac{e^{\|W_{yi}\|\|X_i\|\cos(\theta_{yi})}}{\sum_j e^{\|W_j\|\|X_i\|\cos(\theta_j)}}\right) \quad (2)$$

The purpose of the initial Softmax is to make $W_1^T X > W_2^T X$, that is $\|W_1\|\|X\|\cos\theta_1 > \|W_2\|\|X\|\cos\theta_2$, which gives the correct classification result for sample $X$ (from class 1). The motivation of L-Softmax loss [9] is to generate a decision margin by adding a positive integer variable $m$, which can constrain the above inequalities more strictly. As following:

$$\|W_1\|\|X\|\cos\theta_1 \geq \|W_1\|\|X\|\cos m\theta_1 > \|W_2\|\|X\|\cos\theta_2$$

where $0 \leq \theta_1 \leq \frac{\pi}{m}$.

AM-Softmax [10] rewrites the equation of $\cos(\theta)$ to: $\psi(\theta) = \cos(\theta) - m$. The above formula is simpler than the $\psi(\theta)$ of L-Softmax [9] in form and calculation. In addition, based on L-Softmax [9], a constraint is added: $b = 0$, $\|W\| = 1$. Compared with L-Softmax loss [9], the difference between the classes is only related to the angle of $\theta$, and $m$ is angular margin. So, after the normalization of weights and input features, the loss function can be expressed as:

$$L_{AM} = -\frac{1}{N}\sum_i log\frac{e^{s\cdot(cos\theta_{y_i}-m)}}{e^{s\cdot(cos\theta_{y_i}-m)} + \sum_{j=1,j\neq y_i}^{c} e^{s\cdot cos\theta_j}} \qquad (3)$$

Center loss [11] calculates the class center of several samples of each class in each batch, and then calculates the MSE loss between each sample and the class center.

$$L_C = \frac{1}{2}\sum_{i=1}^{m}\|x_i - c_{y_i}\|^2 \qquad (4)$$

where $c_{y_i}$ represents the center calculated by the $y_i$ class. Finally, the joint loss function is $L = L_{softmax} + L_C$.

In this paper, PEDCC is used to generate an evenly distributed class centroids, replacing the center calculated in Center loss [11], and MSE loss is used to further reduce the distance between the sample and the class center. Secondly, the fixed and evenly distributed PEDCC weights are directly used as the classification layer weights, and are not updated during training. Finally, the two losses are combined and optimized simultaneously, thus achieving the theoretical optimal distribution. We perform feature visualization for the MNIST [12] dataset, and compare various loss functions in Pytorch 1.0 [21] to show the feature distribution in two-dimensional space and three-dimensional space, respectively, with epochs of 30, as shown in Figure 2:

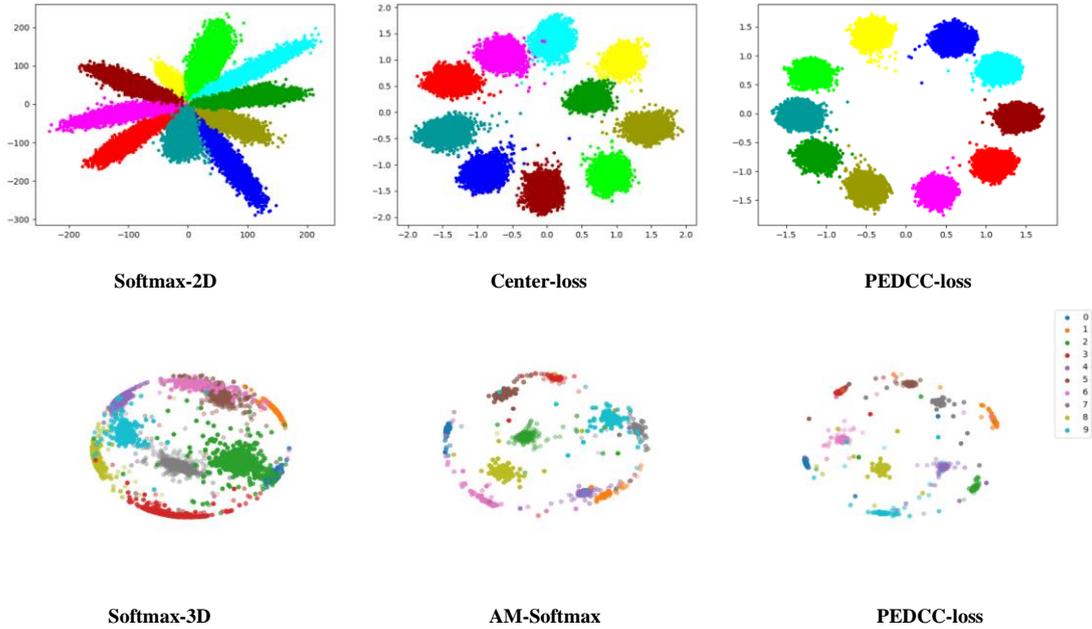

Figure 2 Features visualization of different methods in 2-D and 3-D space

It can be seen that, in the Euclidean space, the PEDCC-loss is distributed on the 2-D or 3-D spherical surface, and each cluster is approximately evenly distributed and compact. While Center loss [11] randomly clustered in the feature space and reduces the intra-class distance, and AM-Softmax [10] has similar result. In the cosine space, PEDCC-loss can not only use margin to separate the inter-class space, but also make the clusters more evenly-distributed, and the sample is closer to the predefined class center.

## 3. Proposed Method

### 3.1 The Distance of Intra-class and Inter-class

From the perspective of statistical pattern recognition and image classification, the original image can be understood as high-dimensional features. Through some traditional machine learning methods, dimensionality reduction is performed on high-dimensional features. The main goal of dimensionality reduction is to generate low-dimensional expressions with higher similarity for intra-class samples and high diversity for inter-class samples, to ensure the compactness of intra-class and the discreteness of inter-class, such as the classic LDA method. Finally, we use the Euclidean distance for image recognition, or the cosine distance for face recognition to classify the samples.

The distance of intra-class is

$$D_{intra} = \sum_{i=1}^{c} P_i (\mu_i - \mu)(\mu_i - \mu)^T \qquad (5)$$

where $\mu_i$ is the center of class $i$, $P_i$ is the probability of class $i$, and $\mu = E[x]$, $\mu_i = E_i[x]$. The distance of inter-class is

$$D_{inter} = \sum_{i=1}^{c} P_i E_i[(\mu_i - \mu)(\mu_i - \mu)^T] \qquad (6)$$

L-Softmax [9] introduced the concept of margin to increase the difficulty of learning, which is more concerned with inter-class distances than traditional Cross Entropy loss. After adding margin, the distribution of each class becomes slender, which improves the gap between classes. In some visual tasks, such as image classification, it also improves the recognition accuracy, but the intra-class distance is not minimal.

Center Loss [11] learns a center for deep features of each class and penalizes the distances between the deep features and their corresponding class centers. Increasing the degree of aggregation within a class is not difficult for a neural network with powerful learning ability. However, Increasing the inter-class distance is a difficult problem. Different classification tasks may have different distances, and the distance between classes is relatively close. If the intra-class distance is large, there will be overlap between class samples. This leads to misclassification, and there is currently no effective way to avoid this problem.

This paper creatively makes use of predefined evenly distribution class centroids, which makes the distance of inter-class be fixed and separated from each other maximally, and simultaneously forces the samples to close to the predefined center as soon as possible.

### 3.2 PEDCC

In this paper, by pre-defining the optimal clustering center, the clustering centers of the classes are artificially set, and these clustering centers are evenly distributed on the hypersphere surface of the feature space, so that the class spacing is maximized.

In this way, we learn a mapping function through CNNs, and map different classes samples to the center of these predefined classes, then to cluster them. So that the distances between different classes can be separated maximally.

The method of generating the predefined class center in this paper is based on the physical model with the lowest isotropic charge energy on the sphere, that is, it is assumed that the $n$ charge points on the hypersphere have repulsive force with each other, and the repulsive force decreases as the distance between the points increases. At the end of the movement, the point on the hypersphere stops moving. Due to the repulsive force, the $n$ points will be evenly distributed on the hypersphere. When the equilibrium state is reached, the $n$ points on the hypersphere can be farthest apart. The detail of algorithm implementation is visible in [13].

As shown in Figure 3, in order to visualize the weight distribution of the PEDCC, we set the output dimensions to 3, and display 4 and 20 classes of PEDCC.

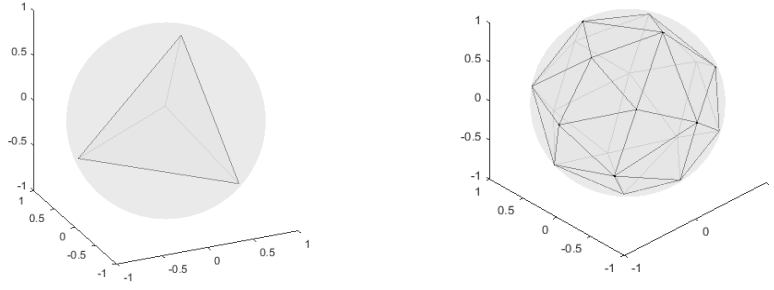

Figure 3 Visualization of the PEDCC weight distribution

### 3.3 PEDCC-Loss

The previous section gives the concept of inter-class distance and intra-class distance in pattern recognition, which is very important in traditional machine learning and deep convolutional neural networks. The essence of machine learning is to learn good feature distribution, and PEDCC gives the theoretically optimal distribution of cluster centers. Therefore, based on the above two concepts, this section will give a new loss function called PEDCC-Loss for CNNs.

The classification layer parameters in the traditional CNNs are trained with the overall network, and the weights are updated using back propagation by minimalize the loss. In the Euclidean space, the score of each sample is calculated by the formula $s_{y_i} = \|W_{y_i}\| \|X_i\| \cos(\theta_{y_i})$. Then we convert scores to probabilities by Softmax function, and get the result of classification.

Because the sample numbers of each class and the quality of the image samples may be different in the dataset, the weight vector $W$ is different too. the visualization of $W$ is the vector from origin to the class center, and the visualization of $X$ is the vector from origin to the point with each different color (corresponding to different classes, see Softmax 2D in Figure 2). Then, the classification layer weight is actually the vector trained by CNNs with sufficient discriminative ability.

The PEDCC is artificially given a plurality of evenly distributed class centers, which are evenly distributed sample points on the unit hypersphere, or a plurality of evenly scattered vectors. Therefore, the global optimal solution of the objective function of the classification layer of CNNs is essentially to obtain a plurality of scattered vectors with sufficient discrimination. We replace the last layer of the convolutional neural network with the predefined class-centered (PEDCC weight), and during the training phase, only the weights of previous layers are updated.

At the end of training phase, in order to obtain better recognition performance, depending on different dataset, a finetuning processing of the PEDCC weight of the last linear classification layer are adopted optionally. PEDCC-loss given as following:

$$L_{PEDCC-AM} = -\frac{1}{N}\sum_i \log \frac{e^{s\cdot(\cos\theta_{y_i}-m)}}{e^{s\cdot(\cos\theta_{y_i}-m)} + \sum_{j=1,j\neq y_i}^{c} e^{s\cdot\cos\theta_j}} \quad (7)$$

$$L_{PEDCC-MSE} = \frac{1}{2}\sum_{i=1}^{m}\|x_i - pedcc_{y_i}\|^2 \quad (8)$$

$$L = L_{PEDCC-AM} + \sqrt[n]{L_{PEDCC-MSE}} \quad (9)$$

Where $s$ and $m$ follow the setting of [20], and $n \geq 1$ is a constrain factor of the PEDCC-MSE. On the unit hypersphere, the distance from the sample to the predefined class center is less than 1, and a nonlinear decision margin *n* is added to the MSE to increase the difficulty of reducing the intra-class distance.

## 4. Experiment Result

### 4.1 Implementation Details

Our experiment is implemented using Pytorch 1.0 [21], which performs image classification and face recognition tasks respectively. Different PEDCC normalized weights are generated according to the number of dataset classes. The network structure of the image classification is the same as [9] where VGG [2] is used, and the batchsize is 256. The network structure of face recognition is the same as [20] where ResNet18 (IRBlock) [3] with 512 features is used, and the batchsize is 128. During the training phase, the initial learning rate is 0.1, the weight attenuation is 0.0005, and the momentum is 0.9. The SGD training algorithm is used for both models.

Since the number of samples in each class in different datasets may be unbalanced, and different classes may have a slightly different clustering propertied, resulting in that a fixed PEDCC weight will not reach the globally optimal state. We allow PEDCC weights to be fine-tuned within a certain range, that is, a PEDCC weight is set a very small learning rate to fine-tune the class center after a certain training epoch. In this paper, the training epochs are 120, so we begin to finetune the PEDCC weight with learning rate 1e-3 at epoch 70 to obtain a globally optimal distribution.

### 4.2 Image Classification Tasks

In the image classification task, the EMNIST [14] data set is first used. The data set has six division methods: ByClass, ByMerge, Balanced, Letters, Digits, and MNIST. We use the Balanced data set for training. The data set has a total of 131,600 characters pictures, and are evenly divided into 47 classes, each class with 2800 characters. The experimental results are shown in Table 1.

Table 1 Accuracy with various loss function in EMNIST

| Loss Function | Accuracy（%） |
|---|---|
| Hinge Loss | 88.22 |
| Cross Entropy Loss | 88.42 |
| L-Softmax (m=2) | 88.69 |
| L-Softmax (m=4) | 88.81 |
| A-Softmax (m=4) | 88.83 |
| Center Loss | 89.21 |
| AM-Softmax (m=0.5) | 89.45 |
| ArcFace (m=0.5) | 89.52 |
| PEDCC-Loss(m=0.5 n=1) | **89.60** |
| PEDCC-Loss -finetuning(m=0.5 n=1) | **89.83** |
| PEDCC-Loss (m=0.5 n=2) | 89.47 |
| PEDCC-Loss -finetuning(m=0.5 n=2) | **89.66** |
| PEDCC-Loss (m=0.5 n=3) | 89.51 |
| PEDCC-Loss -finetuning(m=0.5 n=3) | **89.73** |

Then, we used the more representative CIFAR100 [15] dataset for test, which has 100 natural images, 500 training sets, and 100 test sets. For this dataset, standard data augmentation [9] is performed, that is, the training set image is padding 4 pixels, and then be randomly clipped to 32×32. The 0.5 probability horizontal flip is also performed, while the test set is not processed. The test results are shown in Table 2.

Table 2 Accuracy with various loss function in CIFAR100

| Loss Function | Accuracy（%） |
|---|---|
| Hinge Loss | 67.10 |
| Cross Entropy Loss | 67.26 |
| L-Softmax (m=2) | 70.05 |
| L-Softmax (m=4) | 70.47 |
| A-Softmax (m=4) | 70.86 |
| Center Loss | 71.01 |
| AM-Softmax (m=0.5) | 71.43 |
| ArcFace (m=0.5) | 71.76 |
| PEDCC-Loss (m=0.5 n=1) | **72.22** |
| PEDCC-Loss - finetuning (m=0.5 n=1) | **72.66** |
| PEDCC-Loss (m=0.5 n=2) | **71.87** |
| PEDCC-Loss - finetuning (m=0.5 n=2) | **72.13** |
| PEDCC-Loss (m=0.5 n=3) | 71.59 |
| PEDCC-Loss - finetuning (m=0.5 n=3) | **71.89** |

In CIFAR100 [15], our method predefines 100 classes of 512-dimensional class centers distributed on the hypersphere. After the parameters are solidified, the loss of the training set is also lower than the AM-Softmax [10] of the same parameter. This shows the effectiveness of our method, and the addition of PEDCC-MSE further compresses intra-class distance. In terms of accuracy, PEDCC-loss obtains the best results in the classification.

### 4.3 Face Recognition Tasks

After L-Softamx [9], many studies have focused on the loss function of face recognition, because face recognition pays more attention to the validity of the feature vector, and the increase of the number of classes can better reflect the validity of the loss function. Here we train ResNet18 for the FaceScrub [16] dataset, which contains more than 100,000 face-aligned images for 530 people, with 265 for men and women. After training the model, the 512-dimensional feature vector extracted are used to test the LFW [17] dataset. The training picture size is 144 × 144, which is randomly clipped to 128 × 128, and flipped by the same 0.5 probability level. The number of test faces for LFW [15] is 6000 pairs.

Table 3 Accuracy with various loss function in LFW

| Loss Function | Accuracy（%） |
|---|---|
| Cross Entropy Loss | 91.07. |
| L-Softmax (m=4) | 91.22 |

| A-Softmax (m=4) | 91.74 |
| Center Loss | 92.21 |
| AM-Softmax (m=0.5) | 92.85 |
| ArcFace (m=0.5) | 92.56 |
| PEDCC-Loss (m=0.5 n=1) | 91.43 |
| PEDCC-Loss - finetuning (m=0.5 n=1) | **92.89** |
| PEDCC-Loss (m=0.5 n=2) | 92.04 |
| PEDCC-Loss - finetuning (m=0.5 n=2) | **93.36** |
| PEDCC-Loss (m=0.5 n=3) | 91.27 |
| PEDCC-Loss - finetuning (m=0.5 n=3) | 92.76 |

Through the above experiments, we can know that, compared with the weight of random initialization, the PEDCC weight proposed can get a better weight distribution result and make the model more precise, and a nonlinear factor added to the MSE also can increase the accuracy. Due to the imbalance in the number of samples of various classes, the fixed PEDCC weights are not the optimal state. So, by using the finetuning strategy, we can see that its accuracy has been effectively improved.

## 5. Conclusion

We propose a new loss function based on predefined evenly distributed class centroids for convolutional neural networks. The fixed PEDCC weights are substituted for the parameters of the classification layer in the network, and the improved cross entropy loss is combined with the mean square error of the predefined class center, where a nonlinear factor is also added to the MSE to increase the learning difficulty. Experimental results show that PEDCC-Loss achieves the best results in image classification and face recognition tasks, and network training is stable and easy to converge.